\documentclass[11pt]{article}

\usepackage[final]{acl}

\usepackage{times}
\usepackage{latexsym}
\usepackage[T1]{fontenc}
\usepackage[utf8]{inputenc}
\usepackage{microtype}
\usepackage{inconsolata}
\usepackage{graphicx}
\usepackage{booktabs}
\usepackage{multirow}
\usepackage{amsmath}
\usepackage{amssymb}
\usepackage{xcolor}
\usepackage{booktabs,tabularx,array}

\newif\ifdraft\drafttrue
\ifdraft
  \newcommand{\todo}[2][TODO]{\textcolor{red}{\textbf{[#1: #2]}}}
\else
  \newcommand{\todo}[2][TODO]{}
\fi

\title{Can an LVLM Perceive Event Boundaries Like a Human?}

\author{
Hyolim Kang \quad
Jungsoo Lee \quad
Yeji Song \quad
Whie Jung \quad
Janghoon Cho \quad
Juntae Lee \\
Qualcomm AI Research \\
\texttt{\small \{hyolkang, jungsool, yejisong, whiejung, janghoon, juntlee\}@qti.qualcomm.com}
}

\begin{document}
\maketitle

\begin{abstract}
Large Vision-Language Models (LVLMs) reason effectively about video content, but whether they perceive its event boundaries in a manner consistent with humans remains unclear.
We first investigate this through Generic Event Boundary Detection (GEBD), which identifies event boundaries naturally perceived by humans in continuous video, without relying on predefined event categories.
On Kinetics-GEBD dataset, LVLMs miss many boundaries when predicting directly from frames, but effectively refine candidates proposed by an unsupervised discovery method, achieving state-of-the-art unsupervised GEBD performance.
We extend this discovery-refinement approach to hour-long egocentric video, where boundary discovery requires a broader semantic scale and whole-video LVLM refinement becomes impractical. We address these challenges through coarse-to-fine temporal discovery and localized LVLM accept/reject judgments. Human evaluation shows that the LVLM's agreement with group consensus falls within the range observed among human annotators, with disagreements concentrated on low-consensus cases. Using this pipeline, we construct EgoBoundary, a benchmark for event boundary detection in hour-scale egocentric video. Representative supervised and unsupervised methods degrade substantially on EgoBoundary, highlighting the challenge of long-horizon event boundary detection.
\end{abstract}

\section{Introduction}
\label{sec:intro}
\begin{figure}[t!]
\centering
\includegraphics[width=\columnwidth]{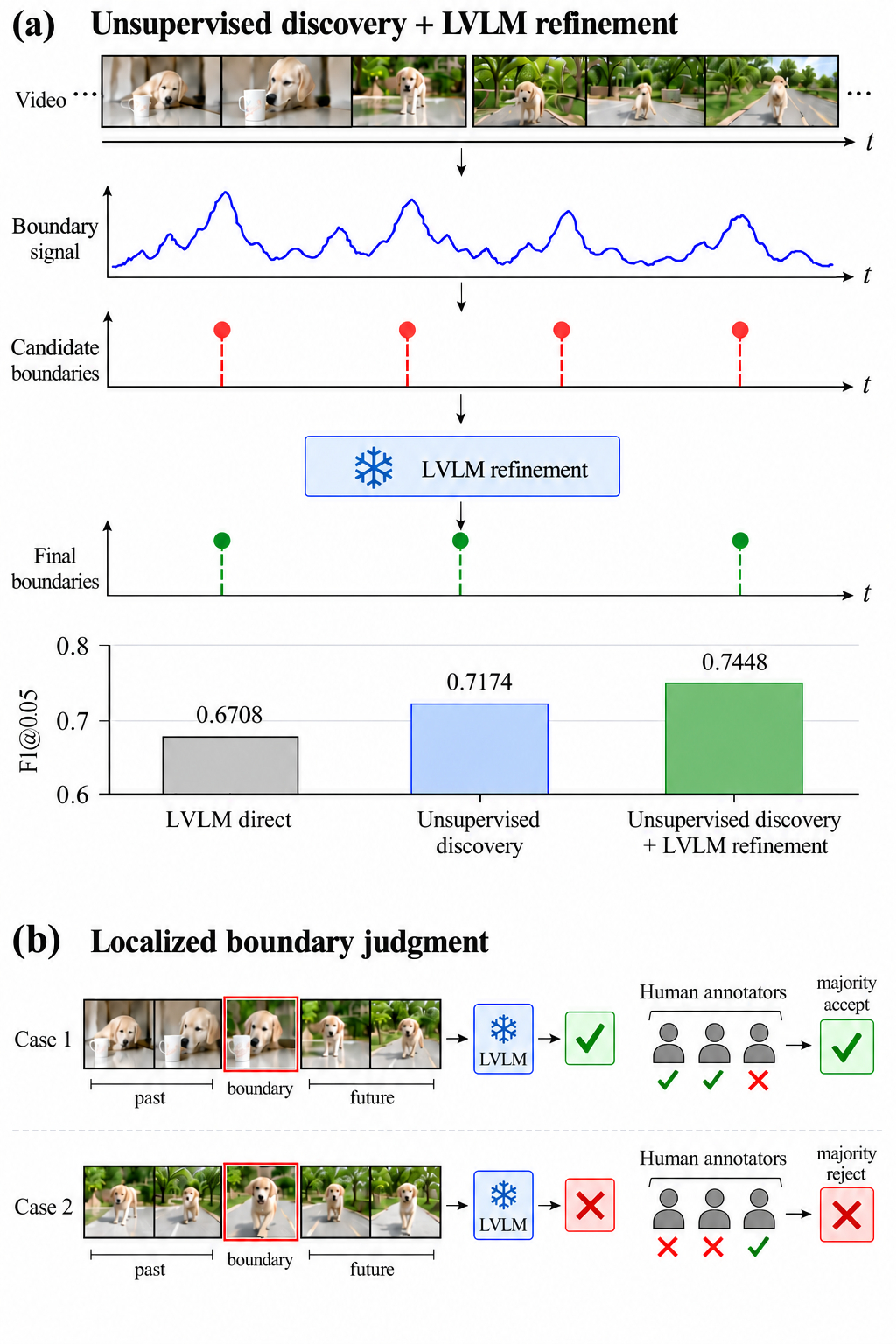}
\caption{Overview of the two key results. (a) Unsupervised discovery followed by LVLM refinement improves event boundary detection on Kinetics-GEBD. (b) Given local visual context around a candidate frame, the LVLM makes accept/reject judgments consistent with human annotators.
}
\label{fig:teaser}
\end{figure}
Humans naturally organize continuous visual experience into meaningful temporal units, perceiving \textit{boundaries} when one coherent event gives way to another~\cite{zacks2007event}.
Generic Event Boundary Detection (GEBD)~\cite{gebd} formalizes this perceptual segmentation without assuming a predefined vocabulary of actions or events: rather than asking what action is occurring, it asks where the perceived structure of an ongoing experience changes. 
As this perceived structure may differ from person to person, event boundaries are inherently subjective. 
GEBD explicitly accounts for this subjectivity by collecting annotations from multiple annotators and using their agreement to identify perceptually salient boundaries.

Recent Large Vision-Language Models (LVLMs)~\cite{qwen2.5vl, gpt5} have demonstrated remarkable capabilities in describing and reasoning about video content, including actions~\cite{llavaction}, interactions~\cite{cott}, and temporal context~\cite{temporalvlm}.
Yet recognizing or describing what happens in a video is different from identifying where humans perceive transitions between meaningful events~\cite{video-llama, video-chatgpt, videochat}.
GEBD provides a natural testbed for this distinction: it requires open-ended semantic judgment over temporal structure, while grounding that judgment in human perceptual consensus.
If general-purpose LVLMs can make such judgments in a human-compatible manner, they could also provide a scalable alternative to costly multi-annotator GEBD annotation, particularly for long, continuous video. 
We therefore ask: \textit{Can an LVLM perceive event boundaries in a manner consistent with humans?}

We first investigate this question on Kinetics-GEBD~\cite{gebd}.
Our experiments reveal a clear asymmetry in what the LVLM can contribute (Fig.~\ref{fig:teaser} (a)).
When asked to detect event boundaries directly from raw video frames, it identifies plausible transitions but systematically misses a substantial portion of the boundary structure.
In contrast, when an unsupervised discovery method first proposes candidate boundaries, the same LVLM becomes an effective refiner, achieving the strongest performance among unsupervised GEBD approaches.
This suggests that current LVLMs benefit from explicit temporal proposals when reasoning about event boundaries, rather than predicting the entire boundary structure directly from raw video.
We therefore adopt a discovery-refinement formulation in which an unsupervised discovery provides candidate coverage and the LVLM contributes semantic judgment.

We next ask whether this discovery-refinement principle can scale to
hour-long untrimmed egocentric video. This setting introduces two challenges. 
First, the appropriate granularity of an event boundary changes with the semantic scale of the video: compared with short Kinetics-GEBD clips, hour-long egocentric streams span much broader activities and contexts, making more macroscopic transitions the natural targets of boundary discovery.
We therefore adopt a coarse-to-fine discovery scheme that identifies transitions at a broader temporal scale and localizes them at a finer one. 
Second, whole-video LVLM refinement becomes impractical at this temporal scale. 
We therefore reformulate refinement as localized accept/reject judgments around individual candidate boundaries (Fig.~\ref{fig:teaser} (b)).

This reformulation turns the LVLM into a local boundary judge, raising a necessary validation question: can such localized LVLM judgments serve as reliable annotations? 
We evaluate the LVLM's accept/reject decisions alongside those of human annotators. 
Its agreement with group consensus falls within the range observed among human annotators, and its disagreements are concentrated on transitions for which humans themselves show weaker consensus. 
These results provide empirical support for using localized LVLM boundary judgments as a scalable annotation mechanism.

Using this pipeline, we construct \textbf{EgoBoundary}, a benchmark for event boundary detection in hour-scale egocentric video.
We evaluate representative supervised and unsupervised GEBD methods and find that existing approaches struggle substantially in this long-horizon setting, exposing the challenge of recovering meaningful temporal structure from continuous egocentric streams.

Our contributions are threefold:
\begin{itemize}
    \item We systematically investigate LVLMs' ability to perceive taxonomy-free event boundaries, revealing that they are more effective at semantic refinement than exhaustive boundary discovery, achieving strong unsupervised performance on Kinetics-GEBD.

    \item We extend this discovery-refinement principle to hour-long egocentric video through multi-scale temporal discovery and localized LVLM boundary judgments, and demonstrate that these judgments align with human perceptual consensus.

    \item We introduce \textbf{EgoBoundary}, a benchmark for event boundary detection in hour-scale egocentric video, and establish representative supervised, training-free, and unsupervised baselines that reveal the limitations of existing methods in long-horizon settings.
\end{itemize}
\section{Related Work}
\label{sec:related}

\subsection{Generic Event Boundary Detection}

Grounded in cognitive studies of event perception~\cite{zacks2007event}, Generic Event Boundary Detection (GEBD)~\cite{gebd} aims to identify moments where humans naturally segment continuous visual experience into meaningful events, without relying on predefined event categories.
As event boundaries are inherently subjective and can occur at different temporal granularities, Kinetics-GEBD guides annotators to identify boundaries at roughly ``one level deeper'' than the video-level event.
Despite this guidance, perceptual differences remain, so Kinetics-GEBD collects multiple annotations per video and retains only videos with sufficient inter-annotator consistency, while preserving diverse valid segmentations.

Building on this benchmark, supervised approaches~\cite{DDMnet, DyBDet, sc-transformer, efficientGEBD, diffgebd} learn discriminative spatiotemporal changes, while unsupervised approaches exploit temporal self-similarity~\cite{kang2021uboco} or optical-flow-based motion cues~\cite{flowgebd} to discover event boundaries.

\noindent\textbf{Our work.}
To quantitatively assess whether LVLMs can perceive event boundaries in a human-consistent manner, we benchmark their GEBD performance against existing methods.
We find that directly prompting LVLMs to predict boundaries from video underperforms existing unsupervised approaches.
In contrast, when candidate boundaries are proposed by an unsupervised temporal signal, LVLMs effectively refine them, substantially outperforming prior unsupervised GEBD methods.

\subsection{LVLMs for Temporal Video Understanding}

Large Vision-Language Models (LVLMs) have demonstrated strong capabilities in understanding and reasoning about video content~\cite{video-llama, video-chatgpt, videochat}.
Recent advances have further improved temporal grounding~\cite{timechat, timesuite, time-r1} and long-video reasoning~\cite{tcot}, enabling models to describe and localize events over extended temporal contexts.
However, understanding what happens in a video does not necessarily translate into the ability to exhaustively identify its temporal structure.

Beyond video understanding, recent studies have explored LVLMs as scalable annotators for temporal localization.
In particular, CoTT~\cite{cott} employs structured LVLM inference to generate dense atomic action annotations from egocentric videos.

\noindent\textbf{Our work.}
We investigate whether LVLMs perceive event boundaries in a manner consistent with humans, beyond action-centric temporal localization.
On Kinetics-GEBD, we quantitatively evaluate how well LVLM-predicted boundaries align with human annotations.
For EgoBoundary construction, we further assess whether localized LVLM accept/reject judgments on candidate boundaries agree with human annotators, validating their potential for scalable annotation of hour-long egocentric video.

\begin{figure*}[t!]
\centering
\includegraphics[width=\textwidth]{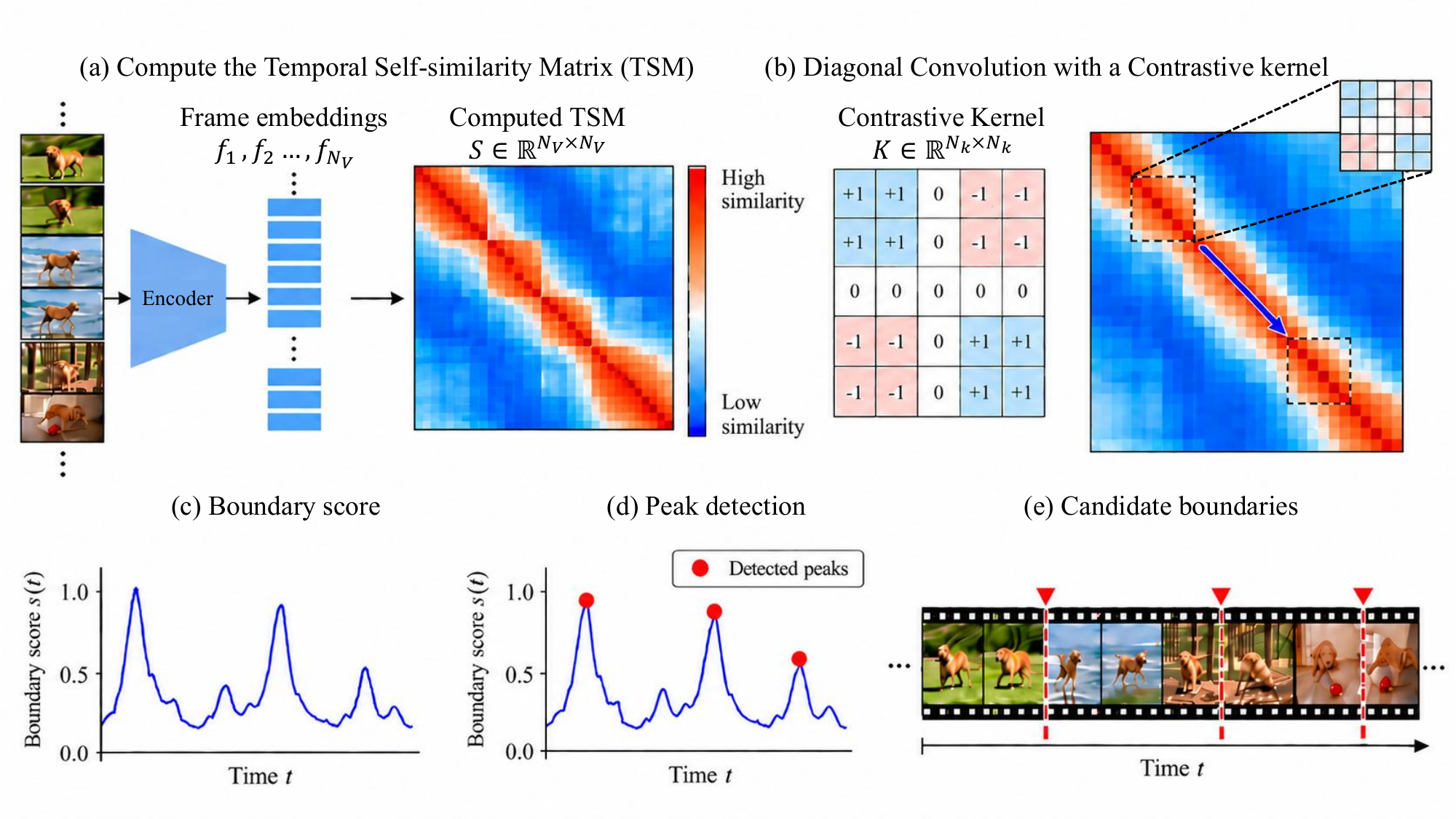}
\caption{Overview of the unsupervised boundary discovery algorithm.
}
\label{fig:unsupervised}
\end{figure*}
\section{Methods}
\label{sec:methods}

\subsection{Problem Definition}
\label{sec:problem_definition}

Generic Event Boundary Detection (GEBD) formalizes the temporal boundaries that humans perceive while segmenting continuous video into meaningful event units. 
Given a video $V$ of duration $T$, let
\begin{equation}
B = \{b_1,\ldots,b_{N_B}\}
\end{equation}
denote a set of reference event boundaries, where $N_B$ is the number of boundaries and each $b_i \in [0,T]$ represents a timestamp perceived as an event boundary.
GEBD aims to predict a boundary set $\hat{B}$ that aligns with these reference boundaries.

\subsection{Probing LVLM Boundary Perception}
\label{sec:lvlm_probe}

We first study how an LVLM can be used for GEBD on short videos. Our goal
at this stage is not to design a specialized boundary detector, but to
probe whether an LVLM can identify human-perceived event transitions and
where its strengths and limitations lie. We consider two inference settings: i) direct LVLM boundary prediction
and ii) two-stage boundary prediction consisting of unsupervised boundary
discovery followed by LVLM refinement.

\paragraph{Probe 1. Direct LVLM inference.}
In the direct setting, we provide the video to the LVLM $\mathcal{F}_{\mathrm{LVLM}}$ and ask it to
identify timestamps at which one coherent event unit transitions to
another. Formally, we write
\begin{equation}
\hat{B}_{\mathrm{direct}}
=
\mathcal{F}_{\mathrm{LVLM}}(V),
\end{equation}
where $\mathcal{F}_{\mathrm{LVLM}}$ directly maps the visual sequence to a
set of predicted event-boundary timestamps. This setting tests whether the
LVLM can both discover candidate temporal changes and determine which of
them constitute meaningful event boundaries without an explicit temporal
proposal mechanism.

\paragraph{Probe 2-1. Unsupervised boundary discovery.}
We next consider a two-stage inference setting, in which candidate boundaries are first discovered using an unsupervised temporal signal and then refined by the LVLM. 
The first stage therefore focuses on identifying potential temporal transitions without relying on LVLM-based reasoning.

As illustrated in Fig.~\ref{fig:unsupervised},
we uniformly sample $N_V$ frames at timestamps
$\{t_i\}_{i=1}^{N_V}$ and extract their
visual embeddings $\{f_i\}_{i=1}^{N_V}$
using a frame encoder~\cite{he2016deep, perceptionencoder}.
We then construct a temporal self-similarity matrix (TSM)
\begin{equation}
S_{ij} = \operatorname{sim}(f_i,f_j),
\end{equation}
where $S \in \mathbb{R}^{N_V \times N_V}$ represents the pairwise
similarity between sampled frames.

Event transitions induce a characteristic local pattern in the TSM:
frames within the same event tend to be more similar, whereas frames across an event transition tend to be less similar~\cite{kang2021uboco}. 
To capture this contrast, we apply a contrastive kernel
$K \in \mathbb{R}^{N_k \times N_k}$, where $N_k$ is odd, along the
diagonal of the TSM. For the $i$-th sampled frame, we compute the boundary
score

\begin{equation}
s(t_i) = \langle K, S^{(i)} \rangle,
\end{equation}
where $S^{(i)}$ is the $N_k \times N_k$ local TSM patch centered at $(i,i)$, and $\langle\cdot,\cdot\rangle$ denotes the element-wise inner product. The
contrastive kernel assigns positive weights to within-event regions and
negative weights to cross-event regions, producing a high response when
the local similarity structure exhibits a boundary-like transition.

Sliding the kernel along the TSM diagonal produces a one-dimensional
boundary-score sequence. We apply peak detection to this sequence and take
the timestamps corresponding to the detected peaks as candidate boundaries:

\begin{equation}
\begin{aligned}
\hat{B}_{\mathrm{cand}}
&=
\left\{
t_i \;\middle|\;
i \in \operatorname{PeakDetect}
\left(\{s(t_j)\}_{j=1}^{N_V}\right)
\right\} \\
&=
\{c_1,\ldots,c_{N_{\mathrm{cand}}}\},
\end{aligned}
\end{equation}
where $N_{cand}$ denotes the number of candidate boundaries.
These candidates provide explicit temporal proposals for the subsequent
LVLM refinement stage.

\paragraph{Probe 2-2. LVLM refinement.}
Rather than requiring the LVLM to search the entire temporal space, we
condition it on the discovered candidates and ask it to refine the proposed
boundary structure:
\begin{equation}
\hat{B}_{\mathrm{refine}}
=
\mathcal{R}_{\mathrm{LVLM}}(V,\hat{B}_{cand}).
\end{equation}
This formulation separates two roles that are entangled in direct
inference: the unsupervised temporal signal is responsible for proposing
potential transition points, while the LVLM is responsible for semantic
judgment over those proposals. Comparing these two settings allows us to
probe whether the LVLM is more effective at exhaustive boundary discovery
or at reasoning about already localized temporal transitions.
\begin{figure}[t!]
\centering
\includegraphics[width=\columnwidth]{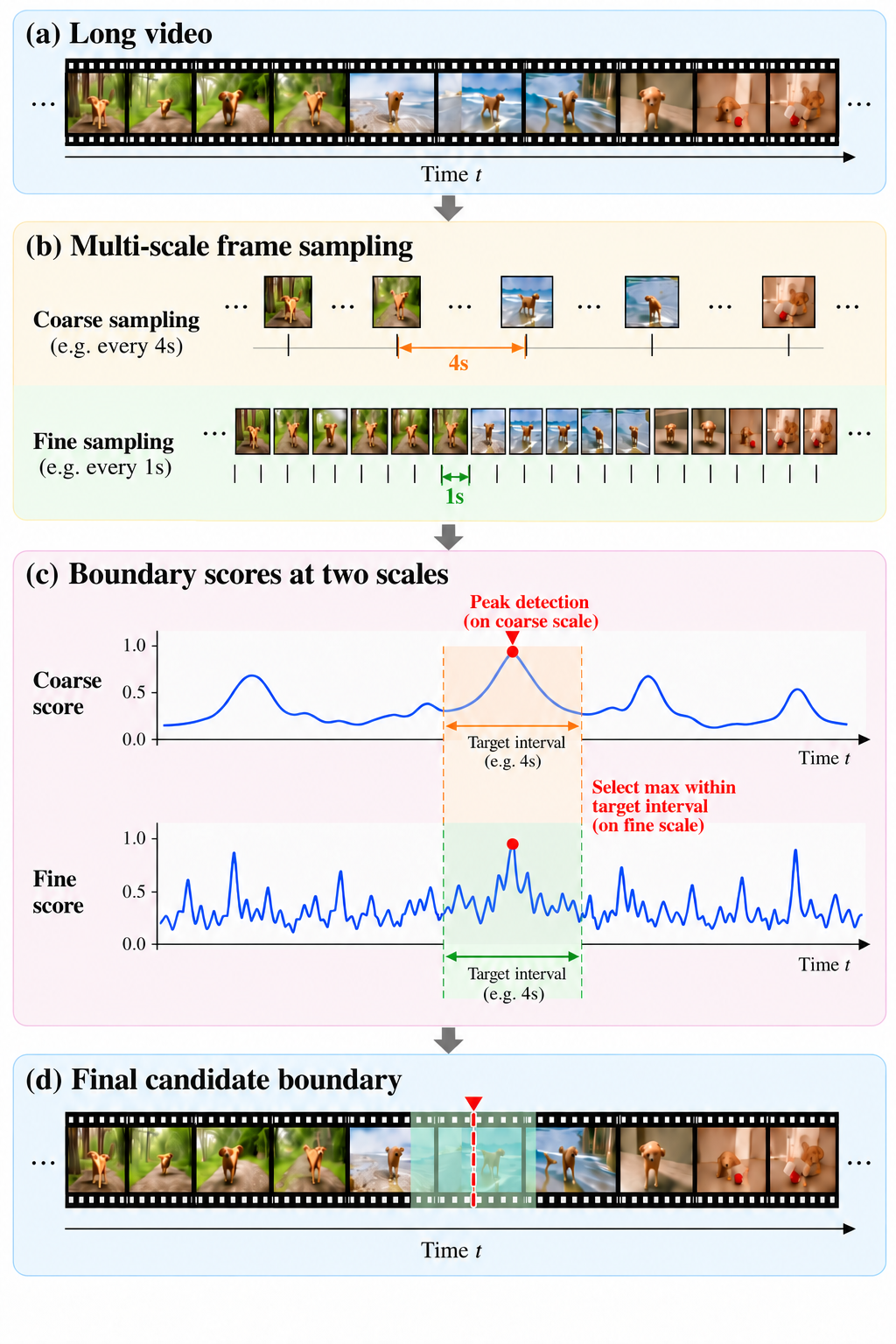}
\caption{Coarse-to-fine discovery}
\label{fig:coarse-to-fine}
\end{figure}

\subsection{Scaling Boundary Perception to Long-Form Egocentric Video}
\label{sec:long_form}

The proposed discovery--refinement formulation provides a natural starting point for
long-form video, but directly applying the short-video refinement procedure
to hour-long egocentric streams is impractical. In particular, repeatedly
reasoning over the full visual sequence requires excessive visual context
and makes temporal localization increasingly difficult. We therefore retain
unsupervised temporal discovery while reformulating LVLM refinement as a
localized boundary judgment.

\paragraph{Coarse-to-fine multi-scale boundary discovery.}
The temporal granularity of an event boundary is relative to the semantic scale of the video. 
While fine-grained transitions are appropriate for short Kinetics-GEBD clips, hour-long egocentric videos span much broader activities and contexts, such that the same ``one-level-deeper'' principle favors more macroscopic event transitions. 
We therefore adopt a coarse-to-fine scheme that discovers boundaries at a broader temporal scale and uses a finer scale only for precise localization.

As illustrated in Fig.~\ref{fig:coarse-to-fine}, we sample the video at two temporal resolutions and compute a boundary-score sequence at each scale using the same TSM-based procedure.
Let $\{t_i^c\}_{i=1}^{N_c}$ and $\{t_j^f\}_{j=1}^{N_f}$ denote the sampled timestamps at the coarse and fine scales, respectively, with corresponding boundary scores $s_c(\cdot)$ and $s_f(\cdot)$.

We perform peak detection only on the coarse-scale
score sequence:
\begin{equation}
P_c = \operatorname{PeakDetect}
\left(\{s_c(t_i^c)\}_{i=1}^{N_c}\right),
\end{equation}
where $P_c$ contains coarse-scale peak indices.

For each coarse peak $p \in P_c$, let $I(p)$
denote the temporal interval associated with
the coarse sample at timestamp $t_p^c$.
We localize the boundary by selecting the
fine-scale timestamp with the highest score
within this interval:
\begin{equation}
c_p =
\underset{t_j^f \in I(p)}{\arg\max}
\; s_f(t_j^f).
\end{equation}
The final candidate set is therefore
\begin{equation}
\hat{B}_{cand}
=
\{c_p \mid p \in P_c\}.
\end{equation}

In this way, the coarse scale determines which temporal transitions are
salient over a broader context, while the fine scale is used only to
localize each selected transition more precisely.

\paragraph{Localized LVLM boundary judgment.}
For each candidate $c_i \in \hat{B}_{cand}$, we extract a local visual
context centered on the candidate and ask the LVLM whether the candidate
corresponds to a meaningful event boundary. We formulate this as a binary
decision
\begin{equation}
y_i
=
\mathcal{J}_{\mathrm{LVLM}}
\left(
V_{c_i}, c_i
\right),
\qquad
y_i \in \{0,1\},
\end{equation}
where $V_{c_i}$ denotes the local visual context around $c_i$, and
$y_i=1$ indicates that the candidate is accepted as an event boundary.
The final prediction is
\begin{equation}
\hat{B}
=
\left\{
c_i \in \hat{B}_{cand}
\mid
y_i = 1
\right\}.
\end{equation}

This localized formulation substantially reduces the visual context presented to the LVLM and converts long-form boundary detection into a sequence of local semantic judgments.

\begin{table*}[t]
\centering
\resizebox{\textwidth}{!}{
\begin{tabular}{c l c c c c c c c c c c c}
\toprule
Supervision & Method
& Rel.Dis=0.05 & 0.1 & 0.15 & 0.2 & 0.25 & 0.3 & 0.35 & 0.4 & 0.45 & 0.5 & Avg \\
\midrule

\multirow{12}{*}{Supervised}
& BMN~\cite{bmn}
& 0.186 & 0.204 & 0.213 & 0.220 & 0.226 & 0.230 & 0.233 & 0.237 & 0.239 & 0.241 & 0.223 \\

& BMN-StartEnd~\cite{bmn}
& 0.491 & 0.589 & 0.627 & 0.648 & 0.660 & 0.668 & 0.674 & 0.678 & 0.681 & 0.683 & 0.640 \\

& TCN-TAPOS~\cite{gebd}
& 0.464 & 0.560 & 0.602 & 0.628 & 0.645 & 0.659 & 0.669 & 0.676 & 0.682 & 0.687 & 0.627 \\

& TCN~\cite{lea2016segmental}
& 0.588 & 0.657 & 0.679 & 0.691 & 0.698 & 0.703 & 0.706 & 0.708 & 0.710 & 0.712 & 0.685 \\

& PC~\cite{gebd}
& 0.625 & 0.758 & 0.804 & 0.829 & 0.844 & 0.853 & 0.859 & 0.864 & 0.867 & 0.870 & 0.817 \\

& Temporal Perceiver~\cite{temporalperceiver}
& 0.748 & 0.828 & 0.852 & 0.866 & 0.874 & 0.879 & 0.883 & 0.887 & 0.890 & 0.892 & 0.860 \\

& SBoCo-Res50~\cite{kang2021uboco}
& 0.732 & - & - & - & - & - & - & - & - & - & 0.866 \\

& DDM-Net~\cite{DDMnet}
& 0.764 & 0.843 & 0.866 & 0.880 & 0.887 & 0.892 & 0.895 & 0.898 & 0.900 & 0.902 & 0.873 \\

& SC-Transformer~\cite{sc-transformer}
& 0.777 & 0.849 & 0.873 & 0.886 & 0.895 & 0.900 & 0.904 & 0.907 & 0.909 & 0.911 & 0.881 \\

& EfficientGEBD~\cite{efficientGEBD}
& 0.783 & 0.851 & - & - & - & 0.901 & - & - & - & 0.913 & 0.883 \\

& LCVSL~\cite{LCVSL}
& 0.768 & 0.848 & 0.872 & 0.885 & 0.892 & 0.896 & 0.899 & 0.901 & 0.903 & 0.906 & 0.877 \\

& DyBDet~\cite{DyBDet}
& \textbf{0.796}
& \textbf{0.858}
& \textbf{0.880}
& \textbf{0.893}
& \textbf{0.901}
& \textbf{0.907}
& \textbf{0.911}
& \textbf{0.915}
& \textbf{0.917}
& \textbf{0.919}
& \textbf{0.890} \\

\midrule

\multirow{7}{*}{Unsupervised}
& SceneDetect~\cite{gebd}
& 0.275 & 0.300 & 0.312 & 0.319 & 0.324 & 0.327 & 0.330 & 0.332 & 0.334 & 0.335 & 0.318 \\

& PA-Random~\cite{gebd}
& 0.336 & 0.435 & 0.484 & 0.512 & 0.529 & 0.541 & 0.548 & 0.554 & 0.558 & 0.561 & 0.506 \\

& PA~\cite{gebd}
& 0.396 & 0.488 & 0.520 & 0.534 & 0.544 & 0.550 & 0.555 & 0.558 & 0.561 & 0.564 & 0.527 \\

& CoSeg~\cite{coseg}
& 0.656 & 0.758 & 0.783 & 0.794 & 0.799 & 0.803 & 0.804 & 0.806 & 0.807 & 0.809 & 0.782 \\

& UBoCo-Res50~\cite{kang2021uboco}
& 0.703 & - & - & - & - & - & - & - & - & - & - \\

& FlowGEBD~\cite{flowgebd}
& 0.713 & 0.828 & 0.850 & 0.858 & 0.862 & 0.864 & 0.866 & 0.867 & 0.868 & 0.869 & 0.845 \\

& \textbf{Ours (Discovery + LVLM)}
& \textbf{0.7448}
& \textbf{0.8367}
& \textbf{0.8612}
& \textbf{0.8738}
& \textbf{0.8810}
& \textbf{0.8855}
& \textbf{0.8888}
& \textbf{0.8916}
& \textbf{0.8940}
& \textbf{0.8965}
& \textbf{0.8654} \\

\bottomrule
\end{tabular}
}
\caption{Comparison with previous methods on Kinetics-GEBD. F1 scores are reported across Rel.Dis. thresholds.}
\label{tab:k400_comparison}
\end{table*}

\section{Experiment}

\subsection{Datasets and Evaluation}
\label{sec:datasets}

\paragraph{Kinetics-GEBD.}
We evaluate on Kinetics-GEBD~\cite{gebd}, a standard GEBD benchmark constructed from short Kinetics-400 clips with multiple independent human annotations per video.
We use only the official validation split, which contains approximately 18K videos.
At Rel.Dis. threshold $d$, a predicted boundary $\hat{b}$ matches an annotated boundary $b$ if
\begin{equation}
\frac{|\hat{b}-b|}{T} \leq d,
\end{equation}
where $T$ is the video duration. Predictions are matched one-to-one against
each annotator independently, and the best-matching annotation is used.
Following the official protocol, we report F1 for $d=0.05$--$0.50$ in
increments of $0.05$, with \textit{F1@0.05 as the primary metric}. To quantify human subjectivity, we additionally divide each video into 20 temporal bins, represent each annotation as binary boundary occupancy, and report the Fleiss' $\kappa$~\cite{fleisskappa} across annotators.

\paragraph{EgoBoundary.}
EgoBoundary is constructed from 42 EgoLife~\cite{egolife} videos totaling 266 hours and contains 5,282 event boundaries produced by our coarse-to-fine discovery and localized LVLM judgment pipeline.
We split the dataset into 35 training and 7 validation videos.
Following established segment-level evaluation protocols in temporal segmentation~\cite{cagqil, actionswitch, openhouse, cott, online-mat, online-hat}, we convert predicted and reference boundaries into temporal segments defined by consecutive boundaries.
We perform one-to-one Hungarian matching between predicted and reference segments based on temporal Intersection over Union (tIoU), counting matched pairs above threshold $\tau$ as correct.
We report segment-level F1 at $\tau \in \{0.5, 0.7, 0.9\}$, adopting $\tau=0.5$ \emph{as the primary metric} following common practice.

\paragraph{Human agreement.}
To validate the localized LVLM judge, we sample 100 candidates judged by Gemini 2.5 Pro, balanced between 50 accepted and 50 rejected cases.
Ten human annotators independently make accept/reject decisions on the same candidate boundaries, using extended local video context. 
We measure inter-annotator agreement among humans using Fleiss' $\kappa$~\cite{fleisskappa}.
To assess individual agreement with human consensus, we compute Cohen's $\kappa$~\cite{cohenkappa} between each judge's decisions and the majority vote of the other human annotators.
For each human annotator, consensus excludes their own judgments (leave-one-out), while the LVLM is evaluated against the consensus of all ten humans.
Tied votes are excluded.

\subsection{Implementation Details}
\label{sec:imp_details}

We use Gemini 2.5 Pro~\cite{gemini} as the LVLM
throughout our experiments.
Details on frame sampling, feature extraction,
boundary discovery, and LVLM refinement are provided
in the supplementary material.

\subsection{Results on Kinetics-GEBD}
\paragraph{Comparison with prior methods.}
Table~\ref{tab:k400_comparison} compares our discovery--refinement approach
with existing GEBD methods on Kinetics-GEBD. Our method achieves an
F1@0.05 of 0.7448, significantly outperforming previous unsupervised approaches at the strictest Rel.Dis. threshold. This advantage is consistently maintained
across the individual Rel.Dis. thresholds, demonstrating that LVLM
refinement effectively improves the temporal boundaries produced by
unsupervised discovery. These results show that combining unsupervised temporal
discovery with LVLM-based refinement provides strong GEBD performance
without supervised training.

\begin{table}[t]
\centering
\resizebox{\columnwidth}{!}{
\begin{tabular}{lccc}
\toprule
Method & Precision & Recall & F1 \\
\midrule
Direct inference
    & \textbf{0.8500} & 0.5540 & 0.6708 \\
Unsupervised discovery
    & 0.6871 & 0.7505 & 0.7174 \\
Combination
    & 0.6910 & \textbf{0.8077} & \textbf{0.7448} \\
\bottomrule
\end{tabular}
}
\caption{Comparison of direct LVLM inference, unsupervised boundary discovery,
and their combination on Kinetics-GEBD at Rel.Dis.~$=0.05$.}
\label{tab:k400_decomposition}
\end{table}
\paragraph{What does the LVLM contribute?}
Table~\ref{tab:k400_decomposition} decomposes the performance of the three
inference settings at the primary Rel.Dis. threshold of $0.05$. Direct LVLM
inference exhibits a strongly precision-oriented behavior, achieving the
highest precision of 0.8500 but substantially lower recall of 0.5540. This
suggests that the LVLM can recognize clear event transitions when it
identifies them, but fails to exhaustively discover the boundary structure
of the video. In contrast, unsupervised discovery provides substantially
higher recall (0.7505), at the cost of lower precision.

Combining the two yields a complementary effect. LVLM refinement significantly increases recall from 0.7505 to 0.8077 while slightly improving precision from 0.6871 to 0.6910, resulting in the best F1 of 0.7448. This corresponds to a
$+0.0274$ improvement over unsupervised discovery alone and a $+0.0740$
improvement over direct LVLM inference. These results indicate that the
LVLM is more effective at refining an explicitly proposed temporal boundary
structure than at discovering all event boundaries directly from the video.

\subsection{Results on EgoBoundary}
\label{sec:ego_results}
Event-boundary perception is inherently subjective even in the standard
Kinetics-GEBD benchmark. Using its original multi-annotator annotations,
we obtain a Fleiss' $\kappa$ of 0.294, indicating substantial
inter-annotator variability. We observe a similar level of disagreement in
our egocentric human study, where the ten annotators achieve a Fleiss'
$\kappa$ of 0.345. Given this inherent subjectivity, we evaluate whether the LVLM's agreement with human consensus falls within the variability observed among individual human annotators, rather than expecting perfect agreement with a single consensus segmentation.
\paragraph{Can the LVLM serve as a local boundary judge?}
Before using localized LVLM judgments to construct EgoBoundary, we first
examine whether these judgments are compatible with human boundary
perception. We compare the LVLM with ten human annotators using
leave-one-out group consensus, treating every judge under the same
evaluation protocol. As shown in Table~\ref{tab:human_ranking}, the LVLM
achieves a Cohen's $\kappa$ of 0.394 and ranks sixth among the eleven
judges. The average human agreement is 0.415, while five human annotators
obtain lower agreement than the LVLM. Thus, despite the inherent ambiguity of event boundary judgments, the LVLM does not behave as an outlier relative to human annotators. Instead, its consistency with human consensus falls within the observed distribution of individual human judgments.

\begin{table}[t]
\centering
\caption{Agreement with leave-one-out group consensus for the LVLM and
human annotators.}
\label{tab:human_ranking}
\small
\setlength{\tabcolsep}{5pt}
\begin{tabular}{c l c}
\toprule
Rank & Judge & $\kappa$ \\
\midrule
1  & Human A & 0.534 \\
2  & Human B & 0.510 \\
3  & Human C & 0.508 \\
4  & Human D & 0.477 \\
5  & Human E & 0.398 \\
\textbf{6}  & \textbf{LVLM} & \textbf{0.394} \\
7  & Human F & 0.388 \\
8  & Human G & 0.377 \\
9  & Human H & 0.369 \\
10 & Human I & 0.337 \\
11 & Human J & 0.248 \\
\bottomrule
\end{tabular}
\end{table}

Importantly, the remaining disagreement is concentrated on boundaries that
humans themselves find ambiguous. We measure human consensus by the vote
margin between accept and reject decisions. Table~\ref{tab:consensus}
shows that, as we restrict evaluation to candidates with stronger human
consensus, the LVLM's agreement increases monotonically from
$\kappa=0.394$ to $0.721$, while the false-reject rate decreases from
41\% to 12\%. This suggests that disagreements are not uniformly
distributed across candidate boundaries, but are concentrated on cases
where human annotators also disagree.

\begin{table}[t]
\centering
\caption{LVLM agreement as a function of human consensus. Margin denotes
the absolute difference between accept and reject votes from ten human
annotators.}
\label{tab:consensus}
\small
\setlength{\tabcolsep}{4pt}
\begin{tabular}{c r c c c}
\toprule
Margin $\geq$ & $N$ & $\kappa$ & Agree. & FR \\
\midrule
2  & 87 & 0.394 & 70\% & 41\% \\
4  & 69 & 0.551 & 78\% & 35\% \\
6  & 58 & 0.597 & 81\% & 33\% \\
8  & 39 & 0.667 & 85\% & 27\% \\
10 & 23 & 0.721 & 87\% & 12\% \\
\bottomrule
\end{tabular}
\end{table}

\paragraph{Does multi-scale discovery produce better candidates?}
We next evaluate the coarse-to-fine discovery strategy used for long-form video.
To isolate candidate quality from candidate quantity, we compare
coarse-to-fine discovery with fine-scale-only discovery under the same
candidate budget of 6,364 proposals. We then apply the same localized LVLM
judge to both candidate sets.

As shown in Table~\ref{tab:multiscale}, coarse-to-fine discovery results in a lower rejection rate of 17.0\%, compared with 22.3\% for fine-scale-only discovery. 
The coarse temporal scale more frequently places proposals at transitions that survive subsequent boundary judgment.

\begin{table}
\centering
\caption{Comparison of coarse-to-fine and single-scale boundary discovery
under the same candidate budget.}
\label{tab:multiscale}
\small
\setlength{\tabcolsep}{5pt}
\begin{tabular}{l r r r}
\toprule
Discovery & Initial & Rejected & Accepted \\
\midrule
Coarse-to-fine
    & 6,364 & 17.0\% & \textbf{83.0\%} \\
Fine only
    & 6,364 & 22.3\% & 77.7\% \\
\bottomrule
\end{tabular}
\end{table}

\begin{table}
\centering
\caption{
Comparison against manually annotated event boundaries.
\#Bnd. denotes the total number of predicted boundaries.
Uniform baselines evenly partition each video using the
same boundary counts as the corresponding predictions
or ground truth.
}
\label{tab:human-eval-hungarian}
\small
\setlength{\tabcolsep}{4pt}
\begin{tabular}{@{}lcccc@{}}
\toprule
Method & \#Bnd. & \multicolumn{3}{c}{F1 @ tIoU} \\
\cmidrule(lr){3-5}
 & & 0.5 & 0.7 & 0.9 \\
\midrule
Discovery
    & 265 & 0.7625 & 0.6188 & 0.4391 \\
+ LVLM refinement
    & 228 & \textbf{0.8276} & \textbf{0.7112} & \textbf{0.5259} \\
\midrule
Uniform ($n$=discovery)
    & 265 & 0.2515 & 0.0639 & 0.0040 \\
Uniform ($n$=refined)
    & 228 & 0.2457 & 0.0474 & 0.0043 \\
Uniform ($n$=GT)
    & 230 & 0.2833 & 0.0687 & 0.0043 \\
\bottomrule
\end{tabular}
\end{table}

\paragraph{Do discovered boundaries align with human perception?}
To further validate our annotation pipeline against human-perceived event boundaries, we manually annotate 3 EgoLife videos totaling 13 hours, yielding 230 reference boundaries.
We evaluate both the initial coarse-to-fine discovery and the LVLM-refined boundaries against these annotations using segment-level tIoU-F1.
To account for the effect of segment density, we additionally consider uniform segmentation baselines.

As shown in Table~\ref{tab:human-eval-hungarian}, unsupervised discovery achieves 0.7625 F1 at tIoU=0.5, far exceeding the count-matched uniform baseline (0.2515), demonstrating its ability to identify meaningful event transitions.
LVLM refinement further improves F1 to 0.8276, with consistent gains at stricter thresholds, demonstrating its effectiveness in filtering spurious boundaries.

\paragraph{How challenging is EgoBoundary?}
Finally, we evaluate representative GEBD approaches on the completed EgoBoundary benchmark.
We consider a supervised method (BasicGEBD~\cite{efficientGEBD}), two unsupervised method (UBoCo~\cite{kang2021uboco}, CoSeg-FFR~\cite{coseg}).
We train BasicGEBD on the EgoBoundary training split and evaluate it on the validation split, while applying UBoCo and CoSeg-FFR without boundary-label supervision.
Table~\ref{tab:ego_baselines} reports segment-level F1 under temporal IoU thresholds of 0.5, 0.7, and 0.9. 
We report both PE-Core feature variants used in our evaluation.

\begin{table}[t]
\centering
\caption{Baseline performance on EgoBoundary using segment-level tIoU-F1.}
\label{tab:ego_baselines}
\small
\setlength{\tabcolsep}{2.5pt}
\begin{tabular}{@{}llcccc@{}}
\toprule
Supervision & Method & Feat. & 0.5 & 0.7 & 0.9 \\
\midrule
\multirow{2}{*}{Supervised}
& \multirow{2}{*}{BasicGEBD}
& G14 & \textbf{0.4783} & \textbf{0.3052} & \textbf{0.1823} \\
& & S16 & 0.4723 & 0.2490 & 0.1307 \\
\midrule
\multirow{4}{*}{Unsupervised}
& \multirow{2}{*}{UBoCo}
& G14 & 0.3120 & 0.1544 & 0.0649 \\
& & S16 & 0.2318 & 0.1181 & 0.0452 \\
\cmidrule(l){2-6}
& \multirow{2}{*}{CoSeg-FFR}
& G14 & 0.2544 & 0.0983 & 0.0133 \\
& & S16 & 0.2720 & 0.1017 & 0.0205 \\
\bottomrule
\end{tabular}
\end{table}

Despite their competitive performance on short (10-second) Kinetics-GEBD clips, existing methods struggle on EgoBoundary, with even the supervised baseline achieving only 0.4783 F1 at tIoU$=0.5$. 
This highlights the distinct challenges of hour-scale egocentric video and leaves substantial room for improvement.

\section{Conclusion}

We show that LVLMs are more effective at refining unsupervised event boundary proposals than discovering them directly. Extending this finding to hour-long egocentric video, we develop a coarse-to-fine discovery and localized judgment pipeline validated against human annotations. We introduce EgoBoundary, a long-horizon GEBD benchmark highlighting the limitations of existing methods.

\section{Limitations}
\paragraph{LVLM coverage and inference cost.}
Our experiments primarily rely on Gemini 2.5 Pro due to the substantial cost of large-scale LVLM inference. Evaluating multiple LVLMs on approximately 18K Kinetics-GEBD videos and 266 hours of egocentric footage would require considerable computational resources. 

\paragraph{Dependence on boundary discovery.}
Our long-form pipeline relies on unsupervised temporal signals to propose candidate boundaries, while the LVLM only accepts or rejects them. Consequently, meaningful transitions missed during discovery cannot be recovered by subsequent refinement. To mitigate this limitation, we use a relatively permissive peak-prominence threshold to favor candidate recall, allowing the LVLM to filter out spurious proposals. 

\paragraph{Scope of human validation.}
Our human agreement study covers 100 candidate boundaries, while independent manual boundary annotations span approximately 13 hours of egocentric video. Although these evaluations support the reliability of our pipeline, they cannot fully capture the subjectivity of event perception across diverse contexts and annotators. Moreover, EgoBoundary is constructed exclusively from EgoLife, leaving generalization to other long-form video domains unexplored.

\bibliography{refs}
\newpage
\appendix
\begin{table}[t]
\centering
\caption{Summary of licenses and usage terms for
external datasets and models.}
\label{tab:licenses}
\small
\setlength{\tabcolsep}{4pt}
\begin{tabularx}{\columnwidth}{
    @{}l>{\raggedright\arraybackslash}X@{}
}
\toprule
Artifact & License / Terms \\
\midrule
Kinetics-GEBD     & CC BY-NC 4.0 \\
EgoLife           & S-Lab License 1.0 \\
Perception Encoder & Apache License 2.0 \\
Gemini 2.5 Pro    & Google Gemini API Terms \\
\bottomrule
\end{tabularx}
\end{table}
\section{License}

Table~\ref{tab:licenses} summarizes the licensing terms
of the key external datasets and models used in this work.
We use external datasets and models for research and evaluation purposes, subject to their respective licensing terms and usage policies.
In particular, the use and redistribution of source videos and derived annotations are subject to the conditions of the original datasets.
Gemini 2.5 Pro is accessed through its API under Google's applicable service terms.

\section{Potential Risk}

EgoBoundary is designed for research on perceptual event segmentation, not for identifying individuals or making consequential decisions.
The source EgoLife videos have faces blurred to protect individual privacy.
We use these videos in accordance with the original dataset's access conditions and usage restrictions.

\section{AI Usage}
We utilized GPT5~\cite{gpt5} to fix typos and polish sentences, and Claude Code for minor code revisions.

\section{Implementation Details}
\label{sec:imp_details}

\paragraph{LVLM access.}
We use a commercial multimodal model (Gemini 2.5 Pro)~\cite{gemini} through its API, supplying sampled video frames. 

\paragraph{Kinetics-GEBD.}
Following the official Kinetics-GEBD protocol, we sample videos at 4 FPS
and use an ImageNet-pretrained ResNet-50 as the visual encoder. Frame
embeddings are used to construct the TSM and compute boundary scores as
described in Sec.~\ref{sec:lvlm_probe}.

\paragraph{EgoBoundary.}
For long-form egocentric videos, we use the coarse-to-fine sampling scheme.
We sample frames at 0.25 FPS for coarse-scale discovery and at 1 FPS for fine-scale localization.
At both scales, we encode frames using the S16 Perception Encoder~\cite{perceptionencoder}.
For peak detection on the coarse-scale boundary scores, we set the peak-prominence threshold such that the resulting candidate boundaries yield an average segment duration of approximately 150 seconds.
For localized LVLM refinement, we provide frames sampled at 1 FPS within a $\pm$20-second window around each candidate boundary and ask the LVLM to accept or reject it.

\section{Prompts and Annotation Instructions}
\label{app:prompts}

This appendix reproduces, in full, every instruction given to a model or to a
human annotator. Section~\ref{app:prompt-k400} covers the Kinetics-GEBD
prompting experiments, Section~\ref{app:prompt-ego} the LLM verification step
used to construct EgoBoundary, and Section~\ref{app:instr-human} the
instructions shown to human annotators.

The Kinetics prompts ask the model to \emph{produce} boundaries, whereas the EgoBoundary prompt asks it to \emph{verify} one proposed boundary at a time and can only remove candidates.

\subsection{Kinetics-GEBD prompting}
\label{app:prompt-k400}

All Kinetics-GEBD prompts address the whole clip at once: the instruction text
is followed by the clip's frames, each preceded by a timestamp line
\verb|[time X.XXs]|, so the model has access to future frames and the setting
is non-causal. Frames are decoded at 4\,fps and center-cropped to
$224\times224$.

\subsubsection{Zero-shot boundary detection}
\label{app:prompt-k400-zs}

{\footnotesize
\begin{verbatim}
You are given frames sampled at a fixed rate in
chronological order from a short video clip. Each
frame is preceded by its timestamp [time X.XXs]
relative to the start of the clip.

Your task is Generic Event Boundary Detection
(GEBD): identify the moments where a human would
naturally perceive the video splitting into
meaningful temporal segments -- WITHOUT any
predefined taxonomy of actions. A boundary
typically corresponds to one of the following (per
Shou et al., 'Generic Event Boundary Detection',
ICCV 2021):
  1) Change of Subject: a new subject
    (person/object) appears or becomes dominant,
    or the previous dominant subject disappears.
  2) Change of Object of Interaction: the subject
    starts interacting with a new object, or
    finishes interacting with the previous one.
  3) Change of Action: the dominant action changes
    (including a sudden change of direction, or
    repetition boundaries of the same action).
  4) Change of Environment: a significant change
    in color/brightness of the scene or the
    dominant subject's surroundings.

Mark boundaries '1 level deeper' than the
whole-clip event -- i.e., sub-events within the
clip, not just the start/end of the clip itself.
Do not report a boundary within the first or last
0.3s of the clip.

Answer ONLY as compact JSON with boundary
timestamps in seconds (floats), in chronological
order: {"boundaries": [<seconds>, ...]}. If there
are no internal boundaries, return {"boundaries":
[]}.
\end{verbatim}
}

\subsubsection{Candidate refinement}
\label{app:prompt-k400-refine}

Here the model receives the boundary candidates of an automatic detector and
may only move or add boundaries. The instruction is followed by a
line beginning \verb|CANDIDATES (seconds):| and then the clip frames.

{\footnotesize
\begin{verbatim}
You are given frames sampled at a fixed rate in
chronological order from a short video clip. Each
frame is preceded by its timestamp [time X.XXs]
relative to the start of the clip.

Your task is Generic Event Boundary Detection
(GEBD): find the moments where a human would
naturally perceive the video splitting into
meaningful temporal segments -- WITHOUT any
predefined taxonomy. A boundary typically
corresponds to (Shou et al., ICCV 2021):
  1) Change of Subject; 2) Change of Object of
    Interaction;
  3) Change of Action (incl. sudden
    direction/repetition change);
  4) Change of Environment
    (color/brightness/surroundings).

An automatic detector has produced a candidate
list of boundary timestamps for this clip (shown
below as CANDIDATES).

Refine the list with EXACTLY three allowed
operations:
  - ADD: scan the clip for any clear perceptual
    boundary NOT already covered by a candidate,
    and add its timestamp. If none are missing,
    add nothing.
  - SNAP: for each candidate, if the actual moment
    of change is slightly before/after it, move it
    to the exact frame of change (typically within
    +/-0.5s). Keep it near its original time.
  - DELETE: remove a candidate that does not mark
    a perceptual boundary at all -- the same
    subject, object of interaction, action and
    environment continue across it. Also remove a
    candidate that marks the SAME transition as
    another candidate you keep (a near-duplicate
    that is too close). Delete only when you are
    confident; when in doubt, keep the candidate.
Every candidate must still be represented in your
output (either at its original time or snapped to
a nearby time). Do not add or move a boundary into
the first or last 0.3s of the clip.

Answer ONLY as compact JSON with the FULL boundary
list = all candidates (snapped as needed) PLUS
your additions, chronological order (seconds,
floats): {"boundaries": [<seconds>, ...]}.
\end{verbatim}
}

\subsection{EgoBoundary construction: LLM boundary verification}
\label{app:prompt-ego}

Each candidate boundary produced by the unsupervised discovery is verified in a separate request.
The model sees a $\pm 20$\,s window around the candidate at 1\,fps (up to 41 frames), each frame preceded by a timestamp line \verb|[time H:MM:SS]|, and with a \verb|PROPOSED BOUNDARY| text marker inserted at the candidate's position. 
The decision is removal-only: accepted boundaries are a subset of the candidates and no new boundaries are created.

{\footnotesize
\begin{verbatim}
You are validating a proposed boundary in a long
egocentric (head-mounted, fisheye) video. Below
are frames in chronological order (1 fps), each
tagged [time H:MM:SS]. A boundary is PROPOSED at
[time 1:00:00] (marked '<<< PROPOSED BOUNDARY
>>>').

Question: Is this a GENERIC EVENT BOUNDARY -- i.e.
is there a meaningful overall change in the
situation across this point (a change in activity,
place, interaction, or context), as opposed to the
same ongoing situation continuing? Accept if there
is a clear overall change; reject if it is the
same situation continuing with no real transition.

Answer ONLY as compact JSON: {"accept":
true|false, "confidence": <0..1>, "justification":
"<one or two sentences on what changes or stays
the same>"}
\end{verbatim}
}

When another candidate boundary already falls inside the same window --- that
is, within $[t-20, t)$ --- a duplicate-suppression cue is appended to the
first paragraph and a \verb|PREV BOUNDARY| marker is
inserted at that earlier position:

{\footnotesize
\begin{verbatim}
NOTE: another boundary was ALREADY placed earlier
within this same window (marked '<<< PREV BOUNDARY
>>>', at 0:59:48). If the change at the PROPOSED
point is essentially the SAME transition already
captured by that previous boundary (i.e. a
near-duplicate that is too close), REJECT it.
Accept only if it marks a genuinely distinct
additional change.
\end{verbatim}
}

\subsection{Human annotation instructions}
\label{app:instr-human}

Annotators judged each candidate boundary from a $\pm 60$\,s video clip
centered on it, played back with speed controls. A red
\verb|BOUNDARY| overlay marks the candidate for about 2.5\,s as playback passes
it; if another candidate lies earlier in the same window, an orange
\verb|PREV BOUNDARY| overlay marks it. Later candidates are deliberately not
shown, so that clusters of nearby candidates are resolved left to right and
exactly one survives per cluster. 

The full text was shown before the task began, and the summary of Section~\ref{app:instr-human-summary} remained on screen throughout.

\begin{figure*}
\centering
\includegraphics[width=\textwidth]{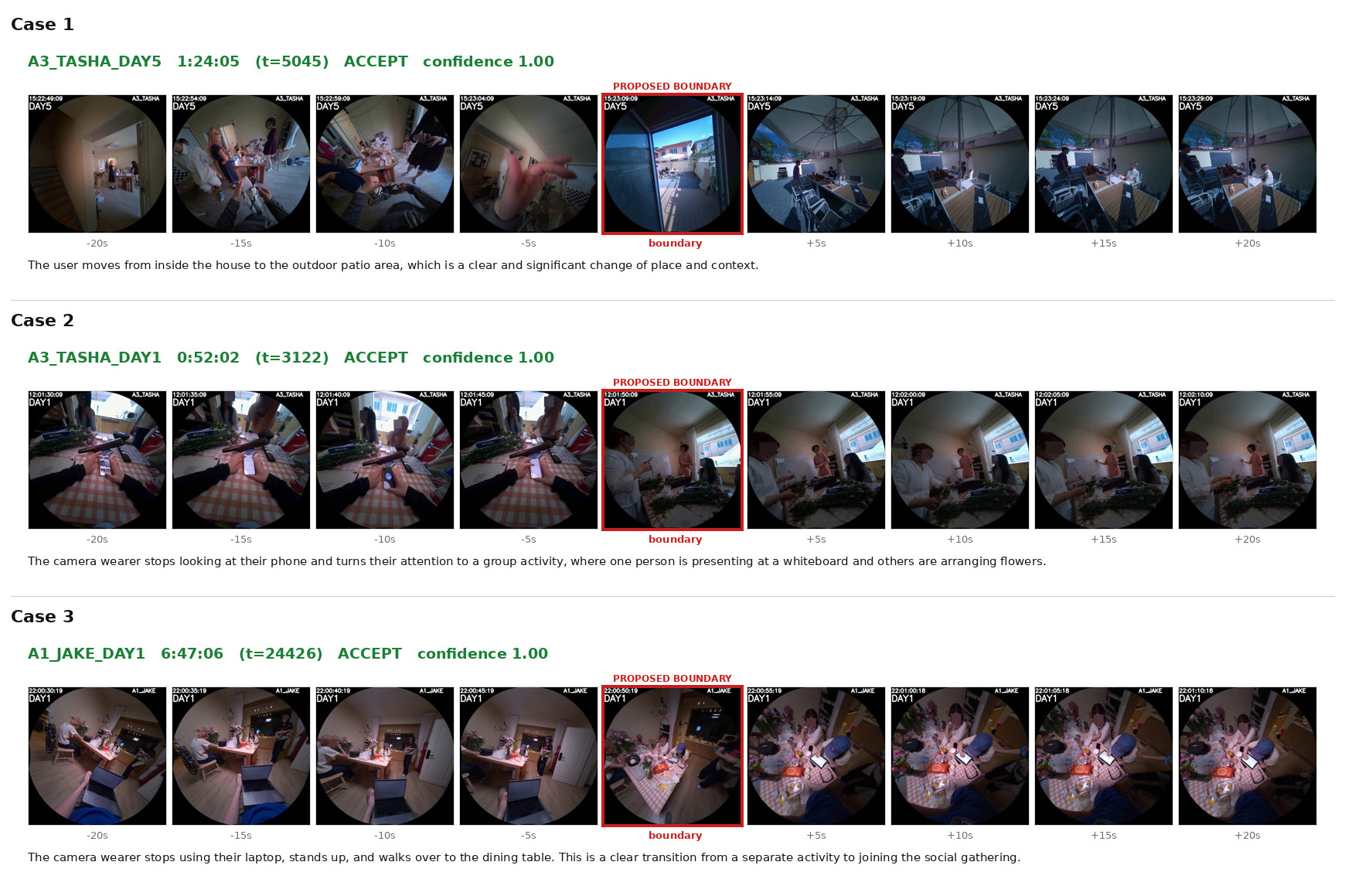}
\caption{
Qualitative examples of \textbf{accepted} event boundaries.
The LVLM identifies meaningful transitions in
location, activity, or social context, including
moving outdoors, shifting attention to a group
activity, and joining a social gathering.
The proposed boundary is highlighted in red.
}
\label{fig:qual_accept}
\end{figure*}

\begin{figure*}
\centering
\includegraphics[width=\textwidth]{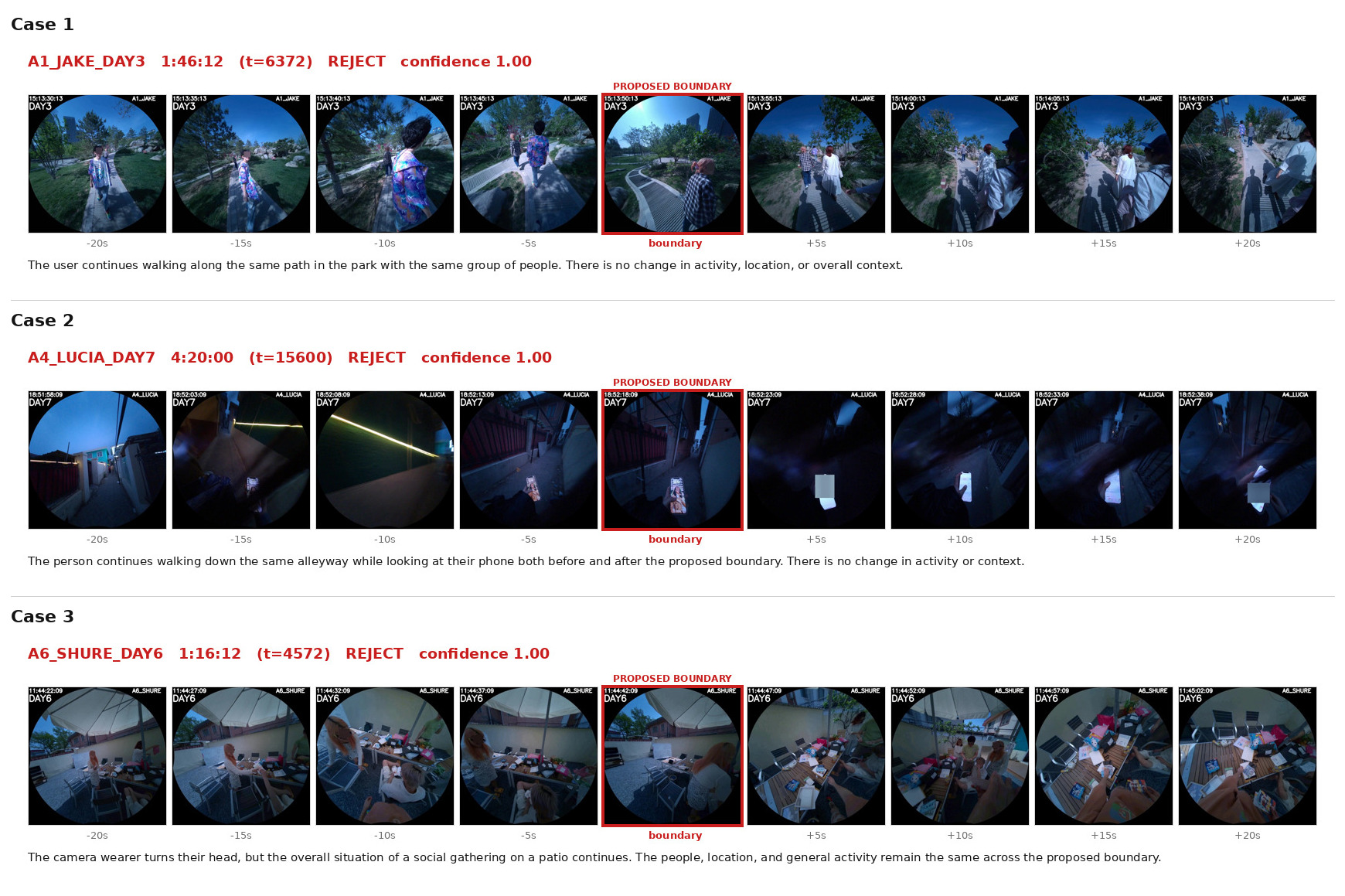}
\caption{
Qualitative examples of \textbf{rejected} event boundaries.
Despite changes in visual appearance or camera
viewpoint, the underlying activity remains
consistent, such as walking along the same path
or continuing a social gathering.
The proposed boundary is highlighted in red.
}
\label{fig:qual_reject}
\end{figure*}

\subsubsection{Full instructions}

\noindent\textbf{Reviewing scene boundaries in video.}
This tool asks you to confirm whether the ``moments where the scene changes''
that were found automatically in first-person (head-mounted camera) daily-life
video are in fact correct. No background in video or AI is needed; judge using
only the criteria below.

\noindent\textbf{What are you judging?} The candidate boundary is the middle of each
clip --- the moment the red \textsc{boundary} marker appears. Looking at that
moment, ask whether the situation (the context) of the scene just before and
the scene just after changes in a meaningful way.

\noindent\textbf{Key question:} ``Across this point, does what the person is doing, or
where they are, or who/what they are interacting with, or the overall
situation, actually change?'' If it changes, \textbf{Accept} (it is a
boundary); if it simply continues, \textbf{Reject} (it is not).

\noindent\textbf{It must be a lasting change, not a momentary one.} Do not look only at
one or two scenes on either side. Ask whether the overall situation itself
changes at that point and whether the new situation then continues for a
while. If the person briefly does something else and soon returns to what they
were doing, that is the same situation continuing, and is not a boundary.

\begin{itemize}
  \item \emph{Not a boundary:} looking at a phone $\rightarrow$ briefly
    glancing up at the surroundings $\rightarrow$ looking at the phone again
    (the ``looking at the phone'' situation is still ongoing).
  \item \emph{A boundary:} looking at a phone $\rightarrow$ putting the phone
    down, moving to another spot, and starting a conversation with people,
    which then continues (the situation changes and persists).
\end{itemize}

There is no exact time criterion. Judge by your own intuition, by whether it
naturally feels like ``the scene moved on here''.

\textbf{Examples of a meaningful change.}
\begin{itemize}
  \item \emph{Change of activity:} cooking $\rightarrow$ starting to eat.
  \item \emph{Change of place:} being in the kitchen $\rightarrow$ moving to
    the living room.
  \item \emph{Change of interaction partner:} looking at a phone alone
    $\rightarrow$ starting a conversation with another person.
  \item \emph{Change of situation or context:} working $\rightarrow$ starting
    to rest.
\end{itemize}
Conversely, if the person keeps doing the same thing, or the camera merely
shakes for a moment, that is not a change: \textbf{Reject}.

\noindent\textbf{When it is too similar to the preceding boundary.} If an orange
\textsc{prev boundary} marker appears earlier in the clip, it means a boundary
was already placed just before this one. If the boundary you are now looking
at is essentially the same change as that preceding one (i.e.\ a duplicate),
reject it as \emph{duplicate of the preceding boundary} (key \texttt{d}). If
it is a different, new change, accept it.

\noindent\textbf{Controls.} \texttt{Space} play/pause; $\leftarrow$ $\rightarrow$ seek
by 2\,s; click the progress bar to seek. \texttt{a} Accept (is a boundary);
\texttt{d} Reject --- duplicate of the preceding boundary; \texttt{x} Reject
--- not a boundary (no change); \texttt{r} Reject --- other;
\texttt{[} \texttt{]} previous/next item. After each judgment the next clip
loads automatically and progress is saved on the server, so the task can be
closed and resumed.

\subsubsection{Summary shown on screen during the task}
\label{app:instr-human-summary}

\textbf{Is this point a generic event boundary?} \quad
\texttt{a} Accept (is a boundary) \quad
\texttt{d} Reject --- duplicate of the preceding boundary (too close) \quad
\texttt{x} Reject --- not a boundary (no transition) \quad
\texttt{r} Reject --- other

\medskip
\noindent\textbf{Criteria (summary).} Across this point, does the situation (the
context) itself change in a meaningful way, and does that change then last for
a while?
\begin{itemize}
  \item \textbf{Accept} --- the activity, place, interaction, or situation
    actually changes and the new situation persists.
  \item \textbf{Reject} --- the same situation continues, or the person
    briefly departs from it and soon returns (e.g.\ looking at a phone,
    glancing up, then back to the phone), or only the camera shakes.
  \item If \textsc{prev boundary} is visible --- if the current boundary is
    essentially the same change as the preceding one, reject it with
    \texttt{d} (duplicate); if it is a new change, accept it.
  \item There is no fixed time criterion. Judge by intuition, by whether it
    naturally feels like ``the scene moved on here''.
\end{itemize}

\subsubsection{Correspondence with the LLM verification prompt}

The human task of Section~\ref{app:instr-human} and the LLM prompt of Section~\ref{app:prompt-ego} ask the same question --- whether activity, place, interaction, or context changes across the point --- and both suppress near-duplicates using a marker for the preceding boundary only.

\section{Qualitative Results}
\label{app:qualitative}

Figures~\ref{fig:qual_accept} and~\ref{fig:qual_reject}
present qualitative examples of localized LVLM
boundary judgments on EgoBoundary.
Each example shows sampled frames surrounding a
proposed boundary, along with the LVLM's decision
and explanation.

As illustrated, the LVLM accepts candidates
associated with meaningful changes in event context,
while rejecting incidental visual changes that do
not indicate a transition between coherent events.
These examples illustrate the role of semantic
judgment in refining unsupervised boundary proposals.

\end{document}